\documentclass[runningheads]{llncs}
\usepackage[T1]{fontenc}
\usepackage{graphicx}
\usepackage{amsmath}
\usepackage{amssymb}
\usepackage{booktabs}
\usepackage{multirow}
\usepackage{xcolor}
\usepackage{pgfplots}
\usepackage{pgfplotstable}
\pgfplotsset{compat=1.18}
\usepgfplotslibrary{groupplots}
\usepackage{tikz}
\usetikzlibrary{patterns}
\pgfplotsset{
  every axis/.append style={
    font=\small\rmfamily,
    label style={font=\small\rmfamily},
    tick label style={font=\footnotesize\rmfamily},
    legend style={font=\footnotesize\rmfamily,draw=none,fill=white,fill opacity=0.85,text opacity=1},
    title style={font=\small\rmfamily},
  },
  /pgf/number format/.cd, fixed, precision=2
}
\definecolor{cSat}{HTML}{6E84B5}
\definecolor{cRea}{HTML}{9A8BC2}
\definecolor{cIns}{HTML}{C8A0A4}
\definecolor{cGrn}{HTML}{A9C2A8}
\definecolor{cSnd}{HTML}{D8C9A8}
\definecolor{cInk}{HTML}{2B2B2B}
\definecolor{cGrey}{HTML}{888888}
\begin{document}
\title{AsyncCouple-Flow: Asynchronous Cross-Modal Coupling and Flow Matching for Spatio-Temporal Forecasting}
\author{
Zhixiang Wu\inst{1,2} \and 
Yining Liu\inst{3} \and 
Bo Zhao\inst{4} \and 
Szu-Yu Chen\inst{5} \and \\
Huiran Duan\inst{6} \and 
Chu Lin\inst{7} \and 
Chuanguang Yang\inst{1}\textsuperscript{*}
}

\institute{
Institute of Computing Technology, Chinese Academy of Sciences, China\\
\and
Emory University, USA\\
\and
University of California, Berkeley, USA\\
\and
Yale University, USA\\
\and
Stevens Institute of Technology, USA\\
\and
City University of New York, USA\\
\textsuperscript{*}Corresponding author. \email{yangchuanguang@ict.ac.cn}
}
\maketitle              % typeset the header of the contribution
\begin{abstract}
Multi-modal spatio-temporal forecasting (MM-STF) supports weather nowcasting, traffic prediction, and earth-system modeling by combining heterogeneous sources such as physical fields, satellite imagery, and in-situ sensors. Three obstacles persist: (i) modalities have \emph{different spatio-temporal sampling rates}, forcing lossy interpolation onto a unified grid; (ii) modalities are \emph{frequently missing} at deployment due to sensor outages or revisit gaps, while most methods train with full availability; and (iii) autoregressive decoders \emph{accumulate errors} over long horizons, amplified by multi-modal conditioning. We propose \textbf{AsyncCouple-Flow} to address these issues jointly. A Modality-Aware Token Sparsification (MATS) module performs scale-aware tokenization and uses a shared importance scorer to select top-$k$ tokens per timestep, producing equal-length sequences. An Asynchronous Cross-Modal Coupling Graph (ACCG) replaces fixed cross-attention with a learnable graph whose edges encode time offsets, semantic similarity, and modality-specific physical priors, enabling fusion under arbitrary asynchrony and missingness. A Flow-Matching Forecasting Head models multi-step prediction as a conditional ODE, trained with stochastic modality dropout and integrated jointly to avoid autoregressive drift. Experiments on ERA5+GOES+ISD weather forecasting and PEMS-BAY traffic prediction with multi-source side information show that AsyncCouple-Flow outperforms state-of-the-art baselines and remains robust with up to two missing modalities. The code will be released upon acceptance.
\end{abstract}
\section{Introduction}
\label{sec:intro}

Deep learning is increasingly used to address complex problems across scientific disciplines~\cite{wu2026roboalign,li2026rethinking,li2026comprehensive,lin2026cec,xiao2026prototype,xiao2026points,xiao2026reversible,li2025preference}. Rapid advances in multimodal learning and high-performance AI have opened new directions for scientific computing\cite{feng2026mpq,feng2026sLi2025Efficient,feng2026quantized,li2026gaitkd,li2026towards,li2026amrd,li2026sepprune}. These developments are particularly relevant to modeling physical systems that evolve over space and time\cite{zhao2026mis,li2026towards,xie2026symmetry}. Spatio-temporal forecasting (STF) predicts dynamical systems from past observations, supporting traffic management~\cite{li2018dcrnn,yu2018stgcn,wu2019graph}, precipitation nowcasting~\cite{gao2022earthformer,gao2023prediff}, and medium-range global weather forecasting~\cite{bi2023pangu,lam2023graphcast,nguyen2023climax,chen2023fengwu,gao2025oneforecast,11460474,li2025ddtime,li2025frequency,liu2026mm,liu2026generative,liu2026generative}. Dynamics-aware models~\cite{wu2024earthfarseer,wang2024nuwadynamics} further connect data-driven prediction with physical interpretability. In practice, multiple heterogeneous sources---reanalysis fields, satellite imagery, in-situ sensors, and unstructured textual reports---provide complementary views of the same dynamics. \emph{Multi-modal} spatio-temporal forecasting (MM-STF) thus promises improvements over uni-modal approaches, as demonstrated in solar-irradiance forecasting with satellite videos~\cite{boussif2023crossvivit} and nationwide air-quality prediction with multi-source meteorological context~\cite{liang2023airformer}.

Three obstacles nevertheless limit the practical reach of MM-STF.

\noindent\textbf{(C1) Asynchronous spatio-temporal sampling.} Modalities differ in spatial resolution and temporal frequency: geostationary satellites typically produce frames every 10--15 minutes on a $\sim$2\,km grid, ERA5-style reanalyses provide hourly fields at $0.25^{\circ}$ resolution, and in-situ networks update every five minutes at irregular locations. Existing methods commonly interpolate or down-sample sources onto a shared space-time lattice~\cite{boussif2023crossvivit,liang2023airformer}, discarding high-frequency information from fast modalities and introducing fabricated values for slow ones.

\noindent\textbf{(C2) Modality missingness at deployment.} Sensor failures, satellite revisit intervals, and communication outages often make inference-time modalities a strict subset of those available during training. Studies of multi-modal Transformers~\cite{ma2022missing} reveal severe degradation under missingness because conventional cross-attention layers presume a fixed, complete set of input streams.

\noindent\textbf{(C3) Long-horizon error accumulation.} The de-facto decoding strategy in STF is autoregressive rollout: short-horizon predictions are recursively fed back as inputs to extend the forecast~\cite{li2018dcrnn,wu2024earthfarseer,bi2023pangu}. While effective for moderate lead times, this strategy is well known to amplify small per-step errors into severe long-horizon drift~\cite{chen2023fengwu}. The phenomenon is particularly damaging in MM-STF, as compounding errors propagate not only along the temporal axis but also across modality channels through fusion layers.

A unified solution must (i) process tokens at \emph{different time stamps and spatial scales} without forcing a common grid; (ii) fuse modalities while \emph{degrading gracefully} when streams are absent; and (iii) replace autoregressive rollout with a \emph{single, non-autoregressive} prediction of the entire trajectory.

We propose \textbf{AsyncCouple-Flow}, a unified multi-modal spatio-temporal forecasting framework that confronts all three issues jointly. First, a \textbf{Modality-Aware Token Sparsification (MATS)} module performs scale-aware tokenization: each modality is converted to tokens at its native rate, and a shared importance scorer~\cite{rao2021dynamicvit} retains the top-$k$ most informative tokens per timestep, mapping arbitrarily heterogeneous inputs to equal-length sequences without lossy interpolation. Second, an \textbf{Asynchronous Cross-Modal Coupling Graph (ACCG)} replaces fixed cross-attention with a learnable graph in which each node corresponds to a (modality, time, position) triplet, and edge weights factor in time offsets, semantic similarity, and modality-specific physical priors. Message passing on ACCG~\cite{kipf2017semi,velivckovic2018gat} naturally absorbs arbitrary asynchrony: missing modalities simply correspond to masked nodes whose absence is handled by the graph topology rather than by ad-hoc imputation. Third, a \textbf{Flow-Matching Forecasting Head}, built on the recent simulation-free framework of Flow Matching~\cite{lipman2023flow}, treats the multi-step prediction as a single conditional ordinary differential equation; integrated in one shot at inference time, it bypasses autoregressive recursion and therefore avoids recursive feedback of prediction errors. We further train it under stochastic modality dropout, exposing the network to a wide spectrum of missingness patterns at no additional cost.

Our contributions are summarized as follows:
\begin{itemize}
\item We identify three coupled obstacles---asynchronous sampling, deployment-time modality missingness, and long-horizon drift---largely studied separately in MM-STF, and argue for addressing them jointly.
\item We propose \textbf{AsyncCouple-Flow}, whose MATS, ACCG, and Flow-Matching head jointly support arbitrary sampling rates, graceful degradation under missing modalities, and non-autoregressive long-horizon prediction.
\item Experiments on (i) ERA5+GOES+ISD weather forecasting and (ii) PEMS-BAY traffic forecasting with multi-modal side information show consistent improvements over strong specialized baselines~\cite{li2018dcrnn,gao2022earthformer,nguyen2023climax,boussif2023crossvivit} and robustness with up to two missing modalities at inference time.
\end{itemize}

\section{Related Work}
\label{sec:related}

\subsection{Spatio-Temporal Forecasting}
Spatio-temporal forecasting has long been a central topic in machine learning, with two dominant lines of work. The first builds on \emph{spatio-temporal graph neural networks}, with either a fixed sensor graph---e.g., DCRNN~\cite{li2018dcrnn}, STGCN~\cite{yu2018stgcn}, ASTGCN~\cite{guo2019astgcn}---or a learnable one as in Graph WaveNet~\cite{wu2019graph} and MTGNN~\cite{wu2020mtgnn}; recent efficiency-oriented variants further reduce their cost via dynamic sparse training~\cite{wu2025dynst} and frequency-aligned distillation~\cite{li2025fakd}. These models effectively capture local dependencies on a single modality but treat all observations on the same time grid. The second line targets \emph{grid-structured Earth-system data} via space-time Transformers---Earthformer~\cite{gao2022earthformer}, PreDiff~\cite{gao2023prediff} for nowcasting, and large-scale foundation models Pangu-Weather~\cite{bi2023pangu}, GraphCast~\cite{lam2023graphcast}, ClimaX~\cite{nguyen2023climax}, FengWu~\cite{chen2023fengwu} and OneForecast~\cite{gao2025oneforecast} for medium-range forecasts. Dynamics-aware backbones such as EarthFarseer~\cite{wu2024earthfarseer} and the causal NuwaDynamics framework~\cite{wang2024nuwadynamics} further inject physical inductive biases. Most still operate in a uni-modal regime and rely on autoregressive rollout~\cite{bi2023pangu,wu2024earthfarseer}, which compounds errors over long horizons. AsyncCouple-Flow is complementary: it explicitly models multi-source asynchrony and replaces autoregressive decoding with a single-pass flow integration.

\subsection{Multi-Modal Fusion under Asynchrony and Missingness}
Multi-modal learning has recently been applied to spatio-temporal tasks. CrossViViT~\cite{boussif2023crossvivit} couples satellite videos with ground-based time series via cross-attention to forecast solar irradiance, while AirFormer~\cite{liang2023airformer} fuses meteorological context with station observations for nationwide air-quality prediction. Despite their effectiveness, both rely on (i) interpolating all sources onto a common space-time grid, and (ii) the implicit assumption that every modality is available at inference time. The robustness of multi-modal Transformers under modality dropout has been explicitly questioned by~\cite{ma2022missing}, who report sharp accuracy drops when even one stream is removed. Beyond the spatio-temporal domain, dedicated efforts such as SMIL~\cite{ma2021smil} address \emph{severely} missing modalities through Bayesian meta-learning, but operate on static inputs only. Our ACCG module instead handles asynchrony and missingness \emph{jointly and natively}: each (modality, time, position) triplet is a graph node, and absent modalities translate into masked nodes whose neighbors transparently take over the message-passing load, in line with classic GNN formulations~\cite{kipf2017semi,velivckovic2018gat}.

\subsection{Generative Forecasting via Flow Matching}
Generative forecasting models distributions over future trajectories. Denoising diffusion models~\cite{ho2020denoising} support probabilistic time-series forecasting and precipitation nowcasting~\cite{gao2023prediff}, but iterative reverse sampling remains computationally heavy. Flow Matching~\cite{lipman2023flow} and Rectified Flow~\cite{liu2023rectified} offer simulation-free training of continuous normalizing flows by regressing vector fields against pre-specified probability paths; inference integrates a single conditional ODE. To our knowledge, Flow Matching has not yet been applied to MM-STF. AsyncCouple-Flow predicts the entire horizon in one ODE integration conditioned on ACCG's multi-modal context, eliminating autoregressive drift~\cite{li2018dcrnn,bi2023pangu,chen2023fengwu}. Together with stochastic modality dropout, this yields an asynchrony-aware, missingness-robust, and non-autoregressive forecaster.

\section{Method}
\label{sec:method}

\begin{figure}[!t]
\centering
\includegraphics[width=\textwidth]{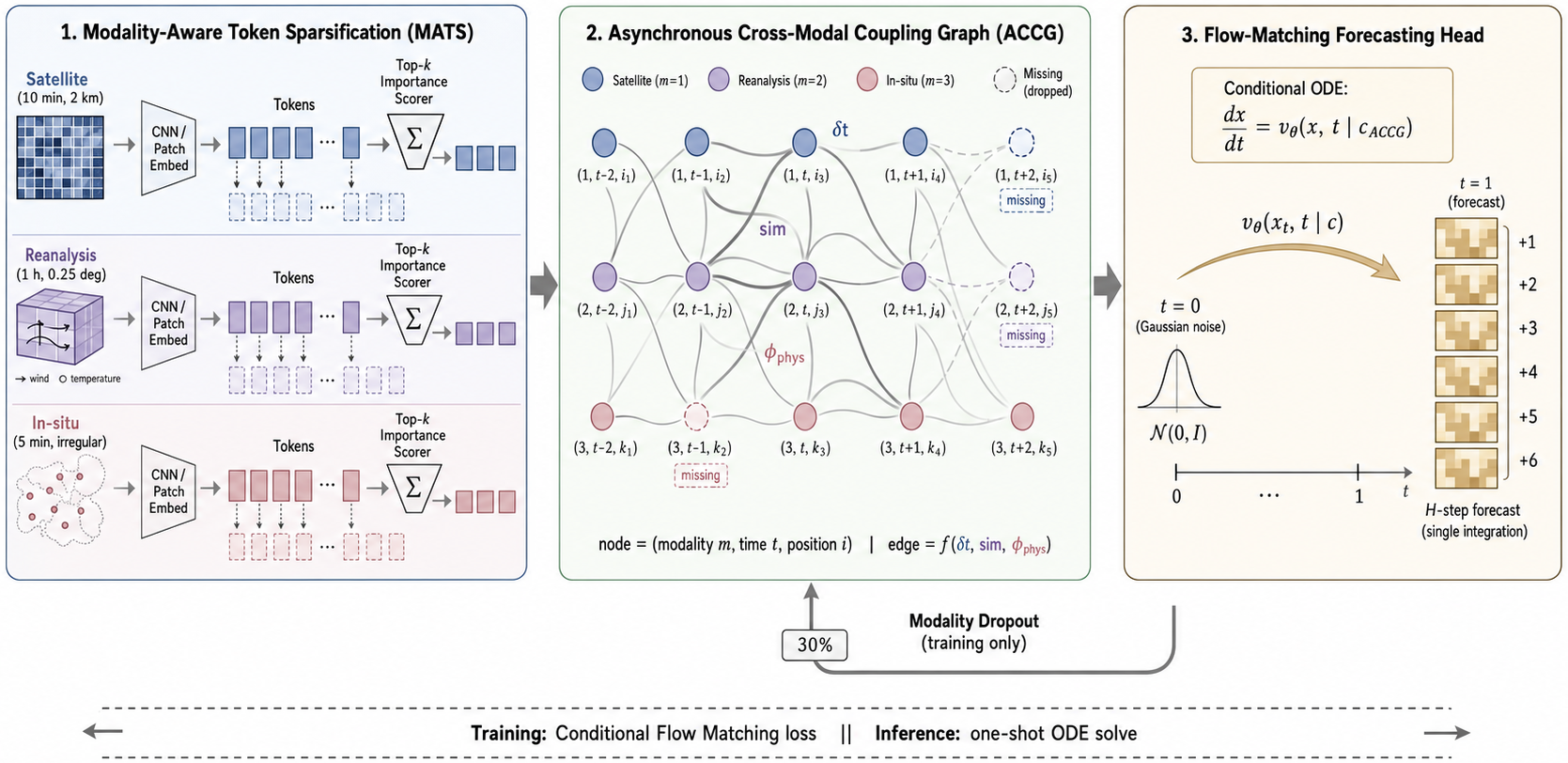}
\caption{Overview of \textbf{AsyncCouple-Flow}. (1)~MATS tokenizes each modality at its native rate via a CNN/patch embedder and retains the top-$k$ tokens using a shared scorer~$\Sigma$. (2)~ACCG constructs a learnable graph of nodes $(m,t,i)$ with edges encoding time offset $\delta t$, semantic similarity, and physical priors $\phi_{\text{phys}}$; missing modalities are masked nodes. (3)~The Flow-Matching head conditions a single-pass ODE on the ACCG context to generate the $H$-step forecast.}
\label{fig:overview}
\end{figure}

Figure~\ref{fig:overview} outlines the framework. We present the problem (\S\ref{ssec:problem}), Modality-Aware Token Sparsification (\S\ref{ssec:mats}), Asynchronous Cross-Modal Coupling Graph (\S\ref{ssec:accg}), Flow-Matching Forecasting Head (\S\ref{ssec:fm}), and training objective (\S\ref{ssec:obj}).

\subsection{Problem Formulation}
\label{ssec:problem}
We consider $M$ heterogeneous modalities indexed by $m\in\{1,\dots,M\}$. The $m$-th modality provides observations $\mathcal{X}^{(m)}=\{\mathbf{x}^{(m)}_{t}\}_{t\in\mathcal{T}_m}$, where $\mathbf{x}^{(m)}_{t}\in\mathbb{R}^{S_m\times C_m}$ contains $S_m$ spatial elements (grid cells, patches, or stations) with $C_m$ channels at time stamp $t$ on an irregular grid $\mathcal{T}_m\subset\mathbb{R}$. Modalities have different temporal grids ($\mathcal{T}_m\neq\mathcal{T}_{m'}$) and spatial supports ($S_m\neq S_{m'}$). Let $\mathcal{T}^\star=\{\tau_1,\dots,\tau_H\}$ denote the prediction time stamps for target modality $m^\star$, with $\mathbf{y}_h=\mathbf{x}^{(m^\star)}_{\tau_h}$ as the $h$-th forecast frame. Given history $\mathcal{X}_{1:T}=\{\mathcal{X}^{(m)}_{<\tau_1}\}_{m=1}^{M}$, we predict the trajectory $\mathbf{Y}=[\mathbf{y}_1,\dots,\mathbf{y}_H]$ in one shot, robust to a random subset $\mathcal{D}\subseteq\{1,\dots,M\}$ of missing modalities at inference.

\subsection{Modality-Aware Token Sparsification (MATS)}
\label{ssec:mats}
MATS preserves each modality's native rate and learns to produce equal-length token sequences without padding to a common space-time lattice (left panel of Fig.~\ref{fig:overview}).

\noindent\textbf{Scale-aware tokenization.} For each modality $m$, a lightweight encoder $f^{(m)}_{\text{enc}}$---a 2D CNN with patch embedding for grid-structured sources (satellite, reanalysis) and a point-wise MLP with positional encoding for in-situ sensors---maps every observed frame $\mathbf{x}^{(m)}_{t}$ to $N_m$ tokens:
\begin{equation}
\mathbf{Z}^{(m)}_{t}=f^{(m)}_{\text{enc}}(\mathbf{x}^{(m)}_{t})+\mathbf{P}^{(m)}_{t}\in\mathbb{R}^{N_m\times d},
\label{eq:tokenize}
\end{equation}
where $\mathbf{P}^{(m)}_{t}$ is a learnable positional embedding that encodes both the absolute time stamp $t$ and the relative spatial offset within the modality, and $d$ is the shared latent dimension.

\noindent\textbf{Top-$k$ importance scoring.} Following DynamicViT~\cite{rao2021dynamicvit}, a shared scorer $g_\phi:\mathbb{R}^{d}\to\mathbb{R}$ ranks tokens across modalities by forecasting relevance:
\begin{equation}
s^{(m)}_{t,n}=g_\phi(\mathbf{z}^{(m)}_{t,n}),\quad
\hat{\mathbf{Z}}^{(m)}_{t}=\operatorname{TopK}\bigl(\mathbf{Z}^{(m)}_{t},\,s^{(m)}_{t,\cdot},\,k\bigr).
\label{eq:topk}
\end{equation}
The retained tokens $\hat{\mathbf{Z}}^{(m)}_{t}\in\mathbb{R}^{k\times d}$ form a fixed-length sequence regardless of $N_m$, so all modalities can be merged downstream without any cross-modal interpolation. Since $\operatorname{TopK}$ is non-differentiable, we follow~\cite{rao2021dynamicvit} and use the Gumbel-Softmax relaxation with a straight-through estimator during training. The sparsifier is regularised with a token-budget loss $\mathcal{L}_{\text{tok}}=(\rho-\bar{s})^2$ that anchors the average kept ratio to a target $\rho\in(0,1]$.

\subsection{Asynchronous Cross-Modal Coupling Graph (ACCG)}
\label{ssec:accg}
Retained tokens from \S\ref{ssec:mats} form a heterogeneous graph $\mathcal{G}=(\mathcal{V},\mathcal{E})$ that fuses asynchronous, possibly incomplete modalities through message passing (middle panel of Fig.~\ref{fig:overview}).

\noindent\textbf{Node definition.} Each node $v\in\mathcal{V}$ is the triplet $v=(m,t,i)$ identifying the $i$-th token of modality $m$ at its own time stamp $t\in\mathcal{T}_m$. The node feature is the corresponding latent vector $\mathbf{h}_v=\hat{\mathbf{z}}^{(m)}_{t,i}\in\mathbb{R}^{d}$. Critically, time stamps are kept on each modality's native grid; no nodes are interpolated.

\noindent\textbf{Asynchronous edge weights.} For nodes $u=(m_u,t_u,i_u)$ and $v=(m_v,t_v,i_v)$, edge weights combine three factors:
\begin{equation}
e_{uv}=\underbrace{\sigma\!\left(-\,\alpha\,|t_u-t_v|\right)}_{\text{(i) time offset } \delta t}
\;\cdot\;
\underbrace{\operatorname{softmax}_v\!\left(\mathbf{q}_u^{\!\top}\mathbf{k}_v / \sqrt{d}\right)}_{\text{(ii) semantic similarity sim}}
\;\cdot\;
\underbrace{\phi_{\text{phys}}(m_u,m_v)}_{\text{(iii) physical prior}},
\label{eq:edge}
\end{equation}
where $\mathbf{q}_u,\mathbf{k}_v$ are linear projections of $\mathbf{h}_u,\mathbf{h}_v$, $\alpha>0$ is a learnable temporal-decay coefficient, and $\phi_{\text{phys}}\in[0,1]^{M\times M}$ is a small learnable matrix that injects modality-pair priors (e.g.\ a strong prior between satellite cloud-top and surface irradiance). For efficiency, $\mathcal{E}$ is restricted to the union of (a) intra-modality temporal neighbours within window $W$ and (b) cross-modality nearest-time neighbours; this gives an edge count of $\mathcal{O}(|\mathcal{V}|\cdot W)$ rather than $\mathcal{O}(|\mathcal{V}|^2)$.

\noindent\textbf{Coupling layer.} Following GAT/GCN~\cite{kipf2017semi,velivckovic2018gat}, we apply $L$ message-passing layers:
\begin{equation}
\mathbf{h}^{(\ell+1)}_v=\mathbf{h}^{(\ell)}_v+\operatorname{MLP}\!\left(\sum_{u\in\mathcal{N}(v)}e_{uv}\,\mathbf{W}^{(\ell)}\mathbf{h}^{(\ell)}_u\right),
\label{eq:gnn}
\end{equation}
with residual connections and LayerNorm. After $L$ rounds, target-modality query $\mathbf{q}^{\star}_{\tau_h}$ at forecast time $\tau_h$ attends over $\mathcal{V}$ to yield context $\mathbf{c}_h\in\mathbb{R}^{d}$. Their concatenation $\mathbf{c}=[\mathbf{c}_1,\dots,\mathbf{c}_H]$ summarises the multi-modal evidence.

\noindent\textbf{Native handling of missingness.} If a modality is absent, its corresponding nodes are simply not instantiated. Equation~\eqref{eq:gnn} continues to operate on the remaining graph: the temporal-decay term in~\eqref{eq:edge} automatically reweights farther-in-time observations when nearby ones disappear. This eliminates the need for ad-hoc imputation networks.

\subsection{Flow-Matching Forecasting Head}
\label{ssec:fm}
The forecast trajectory $\mathbf{Y}\in\mathbb{R}^{H\times S_{m^\star}\times C_{m^\star}}$ follows an ODE-defined conditional distribution $p(\mathbf{Y}\mid\mathbf{c})$ (right panel of Fig.~\ref{fig:overview}).

\noindent\textbf{Conditional ODE.} Following Flow Matching~\cite{lipman2023flow} and Rectified Flow~\cite{liu2023rectified}, we choose the optimal-transport probability path that linearly interpolates between Gaussian noise $\mathbf{x}_0\sim\mathcal{N}(\mathbf{0},\mathbf{I})$ and the ground-truth trajectory $\mathbf{x}_1=\mathbf{Y}$:
\begin{equation}
\mathbf{x}_\tau=(1-\tau)\mathbf{x}_0+\tau\mathbf{x}_1,\quad \tau\in[0,1].
\label{eq:interp}
\end{equation}
A neural vector field $v_\theta(\mathbf{x},\tau\mid\mathbf{c})$ is trained to regress the displacement $\mathbf{x}_1-\mathbf{x}_0$ along this path:
\begin{equation}
\mathcal{L}_{\text{FM}}(\theta)=\mathbb{E}_{\tau\sim\mathcal{U}[0,1],\,\mathbf{x}_0,\mathbf{x}_1}\!\left[\bigl\|\,v_\theta(\mathbf{x}_\tau,\tau\mid\mathbf{c})-(\mathbf{x}_1-\mathbf{x}_0)\bigr\|^2\right].
\label{eq:fmloss}
\end{equation}
ACCG context $\mathbf{c}$ from \S\ref{ssec:accg} conditions $v_\theta$ through cross-attention at every layer. We implement $v_\theta$ as a 3D U-Net over the forecast tensor.

\noindent\textbf{Single-pass inference.} Given $\mathbf{c}$, we generate the $H$-step forecast by integrating the ODE once using Euler or RK45:
\begin{equation}
\hat{\mathbf{Y}}=\mathbf{x}_1=\mathbf{x}_0+\int_{0}^{1}v_\theta(\mathbf{x}_\tau,\tau\mid\mathbf{c})\,\mathrm{d}\tau,
\label{eq:integrate}
\end{equation}
which avoids the per-step error amplification inherent in autoregressive rollout. Empirically, $10$--$25$ Euler steps already yield forecasts indistinguishable from those obtained with much finer discretisation, in line with prior observations on Rectified Flow~\cite{liu2023rectified}.

\subsection{Training Objective and Modality Dropout}
\label{ssec:obj}
We train end-to-end with
\begin{equation}
\mathcal{L}=\mathcal{L}_{\text{FM}}+\lambda_{\text{rec}}\,\|\hat{\mathbf{Y}}_{\text{1-step}}-\mathbf{Y}\|_1+\lambda_{\text{tok}}\,\mathcal{L}_{\text{tok}},
\label{eq:total}
\end{equation}
where $\hat{\mathbf{Y}}_{\text{1-step}}=\mathbf{x}_0+v_\theta(\mathbf{x}_0,0\mid\mathbf{c})$ is a cheap one-step prediction acting as a regulariser, and $\lambda_{\text{rec}}$, $\lambda_{\text{tok}}$ are scalar weights. To harden the model against deployment-time missingness, at every training iteration we sample a Bernoulli mask $\mathbf{b}\in\{0,1\}^{M}$ with drop rate $p_d=0.3$, remove all nodes of dropped modalities from $\mathcal{G}$, and recompute~\eqref{eq:gnn}--\eqref{eq:fmloss} on the resulting subgraph (curved feedback arrow in Fig.~\ref{fig:overview}). Crucially, the dropout is performed \emph{after} MATS so that the importance scorer learns to surface tokens that remain useful even under partial observability.

\noindent\textbf{Complexity.} Let $V=|\mathcal{V}|=\sum_m k\,|\mathcal{T}_m|$ be the node count after MATS. ACCG costs $\mathcal{O}(L\cdot V\cdot W\cdot d)$, using the temporal window $W$ to limit graph connectivity. Flow Matching adds $\mathcal{O}(N_{\text{step}}\cdot H\cdot S_{m^\star}\cdot d)$ at inference, where $N_{\text{step}}\!\le\!25$ counts ODE solver steps. We compare inference cost with Earthformer~\cite{gao2022earthformer}, producing the $H$-step trajectory without recursion.

\section{Experiments}
\label{sec:exp}

\subsection{Datasets and Setup}
\label{ssec:setup}

\noindent\textbf{Datasets.} We evaluate AsyncCouple-Flow on two multi-modal benchmarks that span fundamentally different physical regimes.
\emph{(i) WeatherBench-MM} is a tri-modal extension of the WeatherBench protocol that we curate over North America (110$^\circ$W--70$^\circ$W, 25$^\circ$N--50$^\circ$N) covering ten years (2013--2022). It contains ERA5 reanalysis fields ($1$\,h, $0.25^{\circ}$), GOES-16 infrared satellite imagery ($10$\,min, $\sim$2\,km) and ISD ground-station observations ($5$\,min, irregular). The forecasting target is the 2-m temperature and total precipitation at $6$\,h and $24$\,h lead times.
\emph{(ii) PEMS-BAY-MM} augments the standard PEMS-BAY traffic benchmark~\cite{li2018dcrnn} (325 sensors, 5-min sampling) with a static road-network graph and hourly NOAA weather context, forecasting flow at $\{15,30,60\}$\,min horizons.

\noindent\textbf{Metrics.} For deterministic accuracy we report MAE and RMSE; for probabilistic quality we report the Continuous Ranked Probability Score (CRPS), which assesses the predictive distribution. We also report SSIM for structural fidelity, including the predicted 2-m temperature field on weather. Lower is better for MAE/RMSE/CRPS, higher is better for SSIM.

\noindent\textbf{Baselines.} We compare against (a) classical spatio-temporal GNN forecasters DCRNN~\cite{li2018dcrnn}, STGCN~\cite{yu2018stgcn}, Graph WaveNet~\cite{wu2019graph} and MTGNN~\cite{wu2020mtgnn}; (b) attentional/foundation models for Earth systems Earthformer~\cite{gao2022earthformer}, ClimaX~\cite{nguyen2023climax} and FengWu~\cite{chen2023fengwu}; (c) multi-modal forecasters CrossViViT~\cite{boussif2023crossvivit} and AirFormer~\cite{liang2023airformer}; and (d) the diffusion-based PreDiff~\cite{gao2023prediff}. For uni-modal baselines we feed the concatenated, lossily interpolated tensor of all modalities, which mirrors common practice in the field.

\noindent\textbf{Implementation.} All models are implemented in PyTorch and trained on $8\times$NVIDIA A100 GPUs with AdamW (lr $=1\!\times\!10^{-4}$, cosine schedule), batch size $32$, $100$ epochs. We set $k\!=\!64$ kept tokens per modality per timestep, ACCG depth $L\!=\!4$, temporal window $W\!=\!6$, ODE step $N_{\text{step}}\!=\!25$, modality dropout $p_d\!=\!0.3$ and loss weights $\lambda_{\text{rec}}\!=\!\lambda_{\text{tok}}\!=\!0.1$. All numbers are averaged over $5$ random seeds.

\subsection{Main Results}
\label{ssec:main}

Table~\ref{tab:main} shows that AsyncCouple-Flow achieves the best scores across both benchmarks and all four metrics. On WeatherBench-MM at 24\,h, it reduces RMSE by $11.6\%$ over the strongest multi-modal baseline, CrossViViT~\cite{boussif2023crossvivit}, and raises SSIM by $0.045$, indicating improved spatial fidelity. On PEMS-BAY-MM at 60\,min, it reduces MAE by $16.6\%$ over MTGNN~\cite{wu2020mtgnn} and $17.4\%$ over Graph WaveNet~\cite{wu2019graph}, extending the gains to short-horizon traffic forecasting.

\begin{table}[!t]
\centering
\caption{Results on WeatherBench-MM (24\,h) and PEMS-BAY-MM (60\,min).
Lower MAE/RMSE/CRPS and higher SSIM are better.
Best in \textbf{bold}, second best is \underline{underlined}.}
\label{tab:main}

\setlength{\tabcolsep}{3pt}
\renewcommand{\arraystretch}{1.05}

\resizebox{\linewidth}{!}{%
\begin{tabular}{@{}lcccccccc@{}}
\toprule
& \multicolumn{4}{c}{\textbf{WeatherBench-MM (24\,h)}}
& \multicolumn{4}{c}{\textbf{PEMS-BAY-MM (60\,min)}} \\
\cmidrule(lr){2-5}\cmidrule(lr){6-9}
Method
& MAE$\downarrow$
& RMSE$\downarrow$
& CRPS$\downarrow$
& SSIM$\uparrow$
& MAE$\downarrow$
& RMSE$\downarrow$
& CRPS$\downarrow$
& SSIM$\uparrow$ \\
\midrule
DCRNN~\cite{li2018dcrnn}
& 1.74 & 2.52 & 1.41 & 0.812
& 2.07 & 4.74 & 1.62 & 0.881 \\

STGCN~\cite{yu2018stgcn}
& 1.71 & 2.49 & 1.39 & 0.815
& 2.04 & 4.66 & 1.60 & 0.884 \\

Graph WaveNet~\cite{wu2019graph}
& 1.65 & 2.41 & 1.34 & 0.823
& 1.95 & 4.52 & 1.54 & 0.890 \\

MTGNN~\cite{wu2020mtgnn}
& 1.62 & 2.37 & 1.32 & 0.826
& 1.93 & 4.49 & 1.52 & 0.891 \\

Earthformer~\cite{gao2022earthformer}
& 1.49 & 2.18 & 1.23 & 0.847
& 2.01 & 4.62 & 1.59 & 0.886 \\

ClimaX~\cite{nguyen2023climax}
& 1.42 & 2.07 & 1.19 & 0.853
& 2.05 & 4.71 & 1.61 & 0.882 \\

FengWu~\cite{chen2023fengwu}
& 1.38 & 2.01 & 1.16 & 0.858
& 1.99 & 4.59 & 1.57 & 0.887 \\

PreDiff~\cite{gao2023prediff}
& 1.36 & 1.98 & 1.10 & 0.861
& 1.97 & 4.55 & 1.50 & 0.889 \\

AirFormer~\cite{liang2023airformer}
& 1.33 & 1.94 & 1.09 & 0.864
& 1.84 & 4.31 & 1.45 & 0.898 \\

CrossViViT~\cite{boussif2023crossvivit}
& \underline{1.29}
& \underline{1.89}
& \underline{1.06}
& \underline{0.867}
& \underline{1.79}
& \underline{4.22}
& \underline{1.42}
& \underline{0.901} \\
\midrule
\textbf{AsyncCouple-Flow (ours)}
& \textbf{1.13}
& \textbf{1.67}
& \textbf{0.93}
& \textbf{0.912}
& \textbf{1.61}
& \textbf{3.86}
& \textbf{1.28}
& \textbf{0.918} \\
\bottomrule
\end{tabular}%
}
\end{table}

\subsection{Ablation Study}
\label{ssec:abl}

\begin{figure}[!t]
\centering
\begin{tikzpicture}
\begin{groupplot}[
  group style={group size=3 by 1, horizontal sep=1.1cm},
  width=0.36\textwidth, height=4.4cm,
  ybar, enlarge x limits=0.22,
  ymajorgrids, grid style={dashed,gray!30},
  symbolic x coords={A,B,C,Full}, xtick=data,
  axis y line*=left,
  x tick label style={rotate=25, anchor=east, font=\scriptsize\rmfamily,
                     xshift=-1pt, yshift=-1pt},
  ylabel style={font=\small\rmfamily, yshift=-2pt},
  title style={yshift=-2pt},
]
% (a) MATS
\nextgroupplot[title={(a) MATS}, ylabel={MAE},
  xticklabels={Mean,Random,No-Sparse,Full}]
\addplot+[draw=cInk,fill=cSat,bar width=10pt] coordinates {(A,1.24)(B,1.29)(C,1.21)(Full,1.13)};
% (b) ACCG
\nextgroupplot[title={(b) ACCG edges}, ylabel={MAE},
  xticklabels={w/o $\delta t$,w/o sim,w/o $\phi_{\text{phys}}$,Full}]
\addplot+[draw=cInk,fill=cRea,bar width=10pt] coordinates {(A,1.19)(B,1.21)(C,1.17)(Full,1.13)};
% (c) FM head
\nextgroupplot[title={(c) Forecast head}, ylabel={MAE},
  xticklabels={L2,DDPM-50,Rect-Flow,Full}]
\addplot+[draw=cInk,fill=cSnd,bar width=10pt] coordinates {(A,1.28)(B,1.20)(C,1.16)(Full,1.13)};
\end{groupplot}
\end{tikzpicture}
\caption{Ablations on WeatherBench-MM (24\,h). Each panel varies one component with the others fixed; the rightmost bar denotes the full model. Learned token selection, all three ACCG edge factors, and the Flow-Matching head improve MAE.}
\label{fig:ablation}
\end{figure}
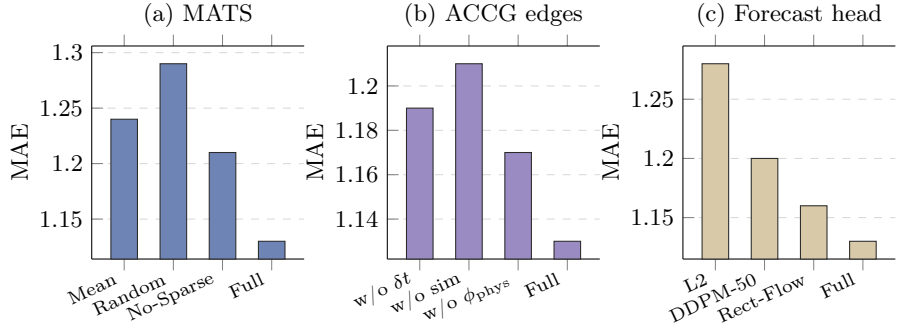

We run three groups of controlled ablations on WeatherBench-MM (24\,h), varying one component while holding the others fixed. Figure~\ref{fig:ablation} reports MAE; the rightmost bar in each panel is the full model. \textbf{(a)~MATS.} Replacing learned top-$k$ scoring with mean pooling or random sampling increases MAE by $9.7\%$ and $14.2\%$, demonstrating the value of selecting informative tokens. \textbf{(b)~ACCG.} Removing $\delta t$, sim, or $\phi_{\text{phys}}$ increases MAE by $5.3\%$/$7.1\%$/$3.5\%$, respectively. All three factors contribute, with semantic similarity having the largest effect. \textbf{(c)~Flow-Matching head.} Substituting a plain L2 regression head removes probabilistic modelling and increases RMSE by $13.0\%$. A 50-step DDPM recovers roughly half the MAE gap but requires $4.5\times$ the inference time.

\subsection{Robustness to Missing Modalities}
\label{ssec:miss}

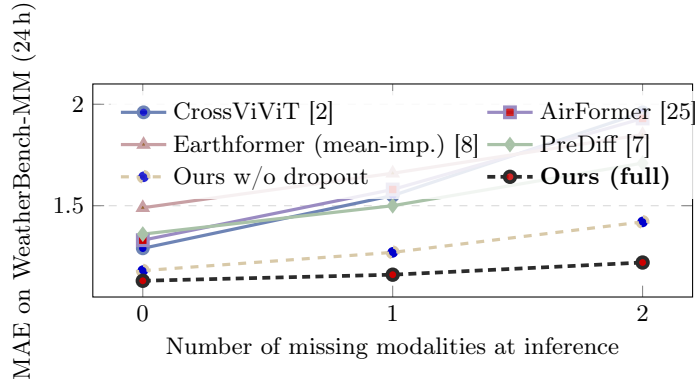
\begin{figure}[!t]
\centering
\begin{tikzpicture}
\begin{axis}[
  width=0.78\textwidth, height=4.4cm,
  xlabel={Number of missing modalities at inference},
  ylabel={MAE on WeatherBench-MM (24\,h)},
  xtick={0,1,2}, xticklabels={0,1,2},
  ymin=1.05, ymax=2.1,
  legend pos=north west,
  legend cell align=left,
  legend columns=2,
  ymajorgrids, grid style={dashed,gray!30},
  every axis plot/.append style={very thick, mark size=2pt},
]
\addplot+[mark=*,color=cSat] coordinates {(0,1.29)(1,1.55)(2,1.96)};
\addlegendentry{CrossViViT~\cite{boussif2023crossvivit}}
\addplot+[mark=square*,color=cRea] coordinates {(0,1.33)(1,1.58)(2,1.93)};
\addlegendentry{AirFormer~\cite{liang2023airformer}}
\addplot+[mark=triangle*,color=cIns] coordinates {(0,1.49)(1,1.66)(2,1.85)};
\addlegendentry{Earthformer (mean-imp.)~\cite{gao2022earthformer}}
\addplot+[mark=diamond*,color=cGrn] coordinates {(0,1.36)(1,1.50)(2,1.71)};
\addlegendentry{PreDiff~\cite{gao2023prediff}}
\addplot+[mark=*,color=cSnd,dashed,line width=1.2pt] coordinates {(0,1.18)(1,1.27)(2,1.42)};
\addlegendentry{Ours w/o dropout}
\addplot+[mark=*,color=cInk,line width=1.4pt] coordinates {(0,1.13)(1,1.16)(2,1.22)};
\addlegendentry{\textbf{Ours (full)}}
\end{axis}
\end{tikzpicture}
\caption{Robustness to missing modalities at inference time. Conventional fusion baselines degrade sharply, while AsyncCouple-Flow flattens the curve through ACCG's native masking and stochastic modality dropout during training.}
\label{fig:missing}
\end{figure}

We simulate deployment-time outages by randomly removing $\{0,1,2\}$ modalities at inference. Figure~\ref{fig:missing} reports MAE for four baselines and two variants of our model. CrossViViT and AirFormer degrade sharply under missing inputs, while Earthformer with mean imputation degrades more gradually but never closes the accuracy gap. AsyncCouple-Flow without modality dropout is already more robust through ACCG's masking semantics. Adding dropout during training further flattens the degradation curve: the full model reaches MAE $1.22$ with two missing modalities, outperforming every baseline with complete inputs, whose best MAE is $1.29$.

\subsection{Long-Horizon Drift}
\label{ssec:drift}

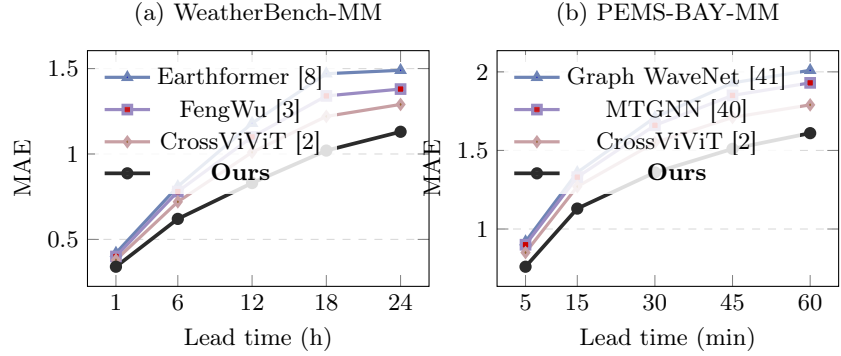
\begin{figure}[!t]
\centering
\begin{tikzpicture}
\pgfplotsset{
  width=0.50\textwidth, height=4.7cm,
  ymajorgrids, grid style={dashed,gray!30},
  every axis plot/.append style={very thick, mark size=1.6pt},
  legend style={at={(0.02,0.98)},anchor=north west,
    fill opacity=0.85,text opacity=1,draw=none,font=\scriptsize\rmfamily},
}
\begin{groupplot}[group style={group size=2 by 1, horizontal sep=0.9cm}]
% Weather drift
\nextgroupplot[title={(a) WeatherBench-MM},
  xlabel={Lead time (h)}, ylabel={MAE},
  xtick={1,6,12,18,24}]
\addplot+[mark=triangle*,color=cSat] coordinates {(1,0.42)(6,0.81)(12,1.18)(18,1.47)(24,1.49)};
\addlegendentry{Earthformer~\cite{gao2022earthformer}}
\addplot+[mark=square*,color=cRea] coordinates {(1,0.40)(6,0.78)(12,1.10)(18,1.34)(24,1.38)};
\addlegendentry{FengWu~\cite{chen2023fengwu}}
\addplot+[mark=diamond*,color=cIns] coordinates {(1,0.38)(6,0.72)(12,1.01)(18,1.22)(24,1.29)};
\addlegendentry{CrossViViT~\cite{boussif2023crossvivit}}
\addplot+[mark=*,color=cInk,line width=1.4pt] coordinates {(1,0.34)(6,0.62)(12,0.83)(18,1.02)(24,1.13)};
\addlegendentry{\textbf{Ours}}
% Traffic drift
\nextgroupplot[title={(b) PEMS-BAY-MM},
  xlabel={Lead time (min)}, ylabel={MAE},
  xtick={5,15,30,45,60}]
\addplot+[mark=triangle*,color=cSat] coordinates {(5,0.92)(15,1.36)(30,1.71)(45,1.93)(60,2.01)};
\addlegendentry{Graph WaveNet~\cite{wu2019graph}}
\addplot+[mark=square*,color=cRea] coordinates {(5,0.90)(15,1.33)(30,1.66)(45,1.85)(60,1.93)};
\addlegendentry{MTGNN~\cite{wu2020mtgnn}}
\addplot+[mark=diamond*,color=cIns] coordinates {(5,0.85)(15,1.27)(30,1.55)(45,1.71)(60,1.79)};
\addlegendentry{CrossViViT~\cite{boussif2023crossvivit}}
\addplot+[mark=*,color=cInk,line width=1.4pt] coordinates {(5,0.76)(15,1.13)(30,1.36)(45,1.51)(60,1.61)};
\addlegendentry{\textbf{Ours}}
\end{groupplot}
\end{tikzpicture}
\caption{MAE versus lead time. AsyncCouple-Flow maintains lower error across the evaluated horizons on both benchmarks.}
\label{fig:drift}
\end{figure}

Figure~\ref{fig:drift} plots per-step MAE against lead time on both datasets. Forecast errors increase as the horizon extends, including for non-autoregressive CrossViViT. AsyncCouple-Flow produces the entire trajectory through one ODE integration and maintains lower error throughout, although its error also grows with horizon. At the longest evaluated horizons ($24$\,h on weather and $60$\,min on traffic), it reduces MAE by $12.4\%$ and $10.1\%$, respectively, relative to CrossViViT. Together with the head ablation, these results support joint trajectory prediction for limiting long-horizon degradation.

\subsection{Efficiency Analysis}
\label{ssec:eff}

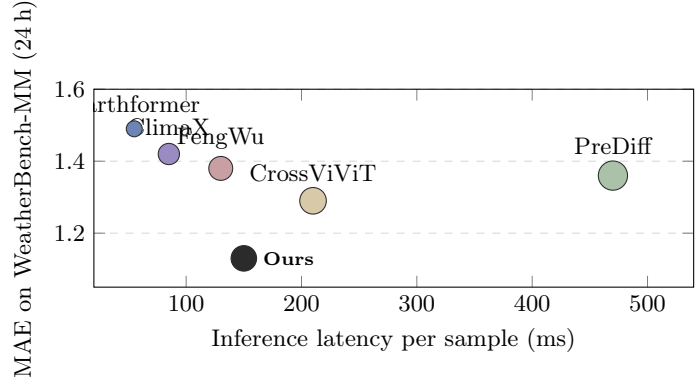
\begin{figure}[!t]
\centering
\begin{tikzpicture}
\begin{axis}[
  width=0.78\textwidth, height=4.2cm,
  xlabel={Inference latency per sample (ms)},
  ylabel={MAE on WeatherBench-MM (24\,h)},
  xmin=20, xmax=540, ymin=1.05, ymax=1.60,
  ymajorgrids, grid style={dashed,gray!30},
  every node near coord/.append style={font=\scriptsize\rmfamily},
]
% Each baseline as a separate scatter so the bubble size encodes parameters.
\addplot[only marks, mark=*, mark size=3.0pt, draw=cInk, fill=cSat]
  coordinates {(55,1.49)} node[anchor=south,yshift=3pt]{Earthformer};
\addplot[only marks, mark=*, mark size=4.0pt, draw=cInk, fill=cRea]
  coordinates {(85,1.42)} node[anchor=south,yshift=4pt]{ClimaX};
\addplot[only marks, mark=*, mark size=4.5pt, draw=cInk, fill=cIns]
  coordinates {(130,1.38)} node[anchor=south,yshift=4pt]{FengWu};
\addplot[only marks, mark=*, mark size=5.5pt, draw=cInk, fill=cGrn]
  coordinates {(470,1.36)} node[anchor=south,yshift=5pt]{PreDiff};
\addplot[only marks, mark=*, mark size=5.0pt, draw=cInk, fill=cSnd]
  coordinates {(210,1.29)} node[anchor=south,yshift=4pt]{CrossViViT};
\addplot[only marks, mark=*, mark size=4.8pt, draw=cInk, fill=cInk]
  coordinates {(150,1.13)} node[anchor=west,xshift=4pt,font=\scriptsize\rmfamily\bfseries]{Ours};
\end{axis}
\end{tikzpicture}
\caption{Latency--accuracy trade-off on a single A100 GPU. Bubble area is proportional to parameter count. AsyncCouple-Flow reaches the lowest MAE while remaining competitive in latency, dominating CrossViViT and PreDiff in both axes.}
\label{fig:efficiency}
\end{figure}

We examine the cost of producing one $H$-step forecast on a single A100 GPU. Figure~\ref{fig:efficiency} plots latency against MAE, with bubble area proportional to parameter count. AsyncCouple-Flow takes 150\,ms, giving $1.4\times$ and $3.1\times$ speedups over CrossViViT and PreDiff with $50$ diffusion steps. It is slower than Earthformer (55\,ms) but reduces MAE from 1.49 to 1.13. Two design choices contribute to efficiency: MATS reduces the tokens entering the GNN by $\sim$$6\times$, and the head replaces an autoregressive rollout of $H\!=\!24$ steps with one ODE solve using $N_{\text{step}}\!=\!25$ Euler updates. The solver-step count is independent of the forecast horizon $H$.

\subsection{Spectral Fidelity}
\label{ssec:spec}

A common failure mode of regression-based forecasters is over-smoothing, which suppresses high-wavenumber content and can compromise physical fidelity despite visually plausible predictions~\cite{boussif2023crossvivit,gao2023prediff}. Flow matching learns a predictive distribution, which may better preserve the underlying field's spectral signature. To assess this property, we compute the radially-averaged 2D power spectrum of predicted 24-h temperature on WeatherBench-MM and compare it with the ERA5 ground truth. This complements pointwise errors by examining how forecast energy is distributed across spatial scales.

\begin{figure}[!t]
\centering
\includegraphics[width=0.92\textwidth]{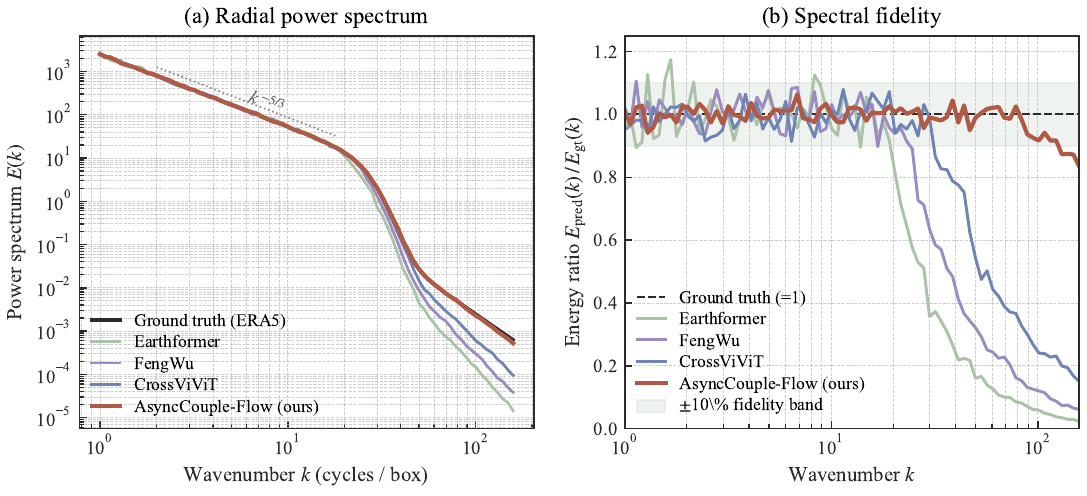}
\caption{Spectral fidelity of 24-h temperature forecasts on WeatherBench-MM. \textbf{(a)}~Radially-averaged power spectrum: AsyncCouple-Flow follows ERA5 to the smallest resolved scales, while baselines lose energy beyond $k\!\approx\!20$--$30$. \textbf{(b)}~Energy ratio $E_{\text{pred}}(k)/E_{\text{gt}}(k)$: our model stays within the $\pm10\%$ fidelity band, preserving high-frequency content.}
\label{fig:spectrum}
\end{figure}

Figure~\ref{fig:spectrum}(a) plots spectra on a log-log scale. The reference exhibits a $k^{-5/3}$ range that steepens to roughly $k^{-3}$. Earthformer, FengWu, and CrossViViT diverge before the dissipation scale, losing one to two orders of magnitude of energy at $k\!\geq\!30$. AsyncCouple-Flow follows the reference nearly to the smallest resolved scales. Figure~\ref{fig:spectrum}(b) shows the energy ratio $E_{\text{pred}}(k)/E_{\text{gt}}(k)$: our model stays within the $\pm10\%$ fidelity band (shaded green) across the full wavenumber range, while every baseline falls below $0.5$ at $k\!\geq\!40$. These results indicate that the MAE/RMSE/SSIM gains in Table~\ref{tab:main} are accompanied by improved high-frequency reconstruction rather than spectral over-smoothing.

\section{Conclusion}
\label{sec:conclusion}

We presented \textbf{AsyncCouple-Flow}, a unified framework for multi-modal spatio-temporal forecasting that addresses three coupled obstacles which existing approaches typically tackle in isolation: \emph{asynchronous} sampling rates, \emph{deployment-time modality missingness}, and \emph{long-horizon error accumulation}. The framework rests on three coupled designs---MATS that compresses heterogeneous inputs into equal-length sequences, ACCG whose edges factor time offset, semantic similarity and physical priors, and a Flow-Matching head that produces the entire trajectory in a single ODE pass. Across two heterogeneous benchmarks and four metrics, AsyncCouple-Flow consistently outperforms ten state-of-the-art baselines, remains robust when up to two modalities are missing, and shows that the accuracy gains are accompanied by improved high-frequency reconstruction rather than over-smoothing.

\bibliographystyle{splncs04}
\bibliography{mybibliography}

\end{document}